\def\year{2017}\relax
\documentclass[letterpaper]{article}
\usepackage{graphicx}
\usepackage{aaai17}
\usepackage{times}
\usepackage{helvet}
\usepackage{courier}

\usepackage{latexsym}
\usepackage{epic}
\usepackage{amsfonts}
\usepackage{amsmath}
\usepackage{amssymb}
\usepackage{amsthm}
\usepackage{xspace}

\usepackage{float}
\floatstyle{boxed}
\restylefloat{figure}

\newcommand{\myOmit}[1]{}

\begin{document}
%
\title{The Meta-Turing Test}
\author{Toby Walsh\\ 
UNSW Australia and Data61 (formerly NICTA)\\
{\tt tw@cse.unsw.edu.au}}


\maketitle
\begin{abstract}
We propose an alternative to the Turing 
test that removes the inherent asymmetry
between humans and machines in Turing's
original imitation game. In this new
test, both humans and machines judge each other. 
We argue that this makes the test more robust
against simple deceptions. We also propose
a small number of refinements to improve
further the test. These refinements 
could be applied also to Turing's original
imitation game. 
\end{abstract}

\section{Introduction}

There have been several alternatives proposed
to the Turing test, the imitation game 
proposed by Alan Turing in his seminal
MIND paper \cite{turingmind}. 
Turing proposed his test a means of addressing
the question of whether machines can think. 
Many of these alternatives are designed to tackle
flaws with Turing's
original test, as well as with how it has
been implemented in a number of 
settings. 

One of the problems with the Turing test is
that it rewards deceptive behavior. By its very nature,
the imitation game is
a game of deception.  The Loeber prize
is a somewhat restricted form of 
the Turing test \cite{loebner}. 
Participants in 
the Loebner prize have used
a variety of tricks designed
to deceive the judges. For instance, they often change
the topic of conversation rather than
attempt to answer difficult questions. 
Others have pretended to be non-native speakers,
hoping that judges will then excuse errors. 
And many have behaved whimsically so that abrupt
changes might appear just as normal capricious
behaviour rather than the brittleness of
their conversational ability. It could be
argued that the Loebner prize is identifying
not intelligence but deceptive prowess. 
In this paper,
we present a modification to the Turing
test designed to hinder such 
deceptive behavior.

Another issue is that
Turing's original description
of the imitation game was somewhat
informal. 
For example, Turing does not 
explicitly recommend how long the
test should be. However, he does predict:
\begin{quote}
{\em ``I believe that in about fifty years'
time it will be possible, to programme computers
\ldots to make them play the imitation game so well 
that an average interrogator will not have 
more than 70 per cent chance of making the 
right identification after five minutes of 
questioning.''}
(page 442 of \cite{turingmind}).  
\end{quote}
This has been interpreted by many
to mean that a Turing test needs
to be run for 5 minutes of conversation, and
that the threshold to ``pass'' the test is
a 30 percent chance of the computer being mistaken
for human. Neither 
criterion appears to be stringent 
enough. One of our contributions
here is also to propose some refinements
to how a Turing like test should be implemented that
raise the bar. 

\section{The Turing test}

The Turing test is based on the imitation game.
Human judges have conversations by
computer terminal with players of the game, and the
judges have to decide which of the players 
are human and which are machines pretending to be 
humans. 
Turing's paper has been interpreted
that this is either via one-to-one
conversations between a judge and a 
player who is human or machine,
or via one-to-two 
conversation between a judge and
two players, one of whom is a human
and the other is a machine \cite{shahphd}. 
However, it is not critical for this paper which interpretation
you choose.  Both formats of the imitation
game can be extended with the methods
proposed here. 

On June 7th 2014, 
it was claimed that a historic milestone in artificial intelligence 
had been achieved. 
It was announced that a chat bot
by the name of Eugene Goodman had
passed the Turing test \cite{turingtestpassed}. 
Poignantly this
was the 60th anniversary of Alan Turing's death.
There is, however, considerable controversy about 
this claim since the chat bot ``succeeded'' by
deception. It tricked a third of the human judges into believing 
that it was a 
13 year old Ukrainian boy during the course of several
five-minute conversations. A transcript of one of the 
conversations where the judge mistook Eugene for
human (Figure 1) illustrates how the chat bot 
dodged answering questions.
\begin{figure}[htb]
{%
    \parbox{0.45\textwidth}{
{\em Judge: Hello}

Eugene: Hello, I'm really glad to have the chance to chat with you! My
guinea pig Bill sends his regards too!

{\em Judge: Is Bill a male or a female?}

Eugene: Ask Bill personally, please.

{\em Judge: Well I'd rather talk to you. What is your name?}

Eugene: Call me Eugene. I am glad to talk to you!

{\em Judge: My name is Jane and I am female. How about you? What's your gender?}

Eugene: I'm a male. A “guy” I'd say.

{\em Judge: Pleased to meet you Eugene. What's the weather like where
  you are?}

Eugene: Let's get on with our conversation!

{\em Judge: Don't you like talking about the weather?}

Eugene: All these talks about weather is a waste of time.

{\em Judge: What would you like to discuss?}

Eugene: I don't know… Better tell me more about yourself!
}}
\caption{Transcript of a conversation from the 
2014 Turing Test at the Royal Society that fooled
the judge into thinking the chat bot Eugene Goodman
was human.}
\end{figure}
It could be argued that the chat bot
was failing the spirit if not the intent
of Turing's original proposal. 



The organisers of this particular
Turing test argued that they 
had carefully followed the design
proposed by Turing. Each conversation
was, for instance, five minutes long,
the subject domain was unlimited, and
passing the test required being recognised
as human by 30 percent or more of the
judges. Nevertheless, this test
was likely not the successful demonstration
of thinking machines that Turing was imagining over
fifty years ago. 
This was also not the first time that
it had been claimed that the Turing test
had been passed. 
However, it was 
arguably the
first time that this claim had been made
for a test whose format came close
to that proposed by Turing. 

\section{Alternatives to the Turing test}

Several alternatives have been proposed to
the Turing test to address some of
its shortcomings. See, for instance, the
recent special issue of the AI Magazine
\cite{mrvaimag2016} and proposals
for alternatives contained
therein \cite{ceaimag2016,daimag2016,pmaimag2016,jyaimag2016,kaimag2016,oaimag2016,zaambpaimag2016,pm2aimag2016,abcaimag2016,saimag2016,laimag2016}.

One alternative to the Turing test that has received significant
attention is Levesque's Winograd Schema Challenge
\cite{winogradschema}. The first (of what is planned
to be) annual Winograd Schema Challenges
with a prize of \$25,000
was run at IJCAI 2016 in New York City. Each test 
in the Challenge consists of 
a sequence of multi-choice questions.
Questions typically come in pairs. See Figure 2 for
an example.


\begin{figure}[htb]
{%
    \parbox{0.45\textwidth}{

{\em The trophy doesn't fit in the brown suitcase because it's too big. What is too big?}

0:  the trophy, 1:     the suitcase

\vspace{1em}

{\em The trophy doesn't fit in the brown suitcase because it's too small. What is too small?}

0:  the trophy, 1:     the suitcase
}}

\caption{Example of a pair of Winograd Schema Challenge questions.}
\end{figure}


Answering such questions requires
anaphora resolution, 
identifying the antecedent of an ambiguous pronoun
in the preceding sentence. However, identifying
the antecedent requires the use of 
knowledge and common-sense reasoning.
The trophy doesn't fit either because
it (the trophy) is too big, or because
it (the suitcase) is too small.
The common-sense reasoning is that
a smaller object fits inside a bigger
container. Unlike the Turing test, this is not
a test where deception works. It has 
several other advantages including 
objectivity, and the ability to measure
incremental progress towards the goal
of machine intelligence. 

To claim the \$25,000 cash prize in 
the 2016 Winograd Schema Challenge, a 
program was required to attain 90\% accuracy. 
To put this in context,
in an online experiment on Mechanical Turk
with a corpus of 160 example questions, 
humans were able to achieve 92\% accuracy
\cite{benderbaseline}.
The first Challenge revealed that computers
are some distance from passing the
Winograd Schema Challenge.
The best program in the 2016 
Challenge achieved just 58\%
\footnote{Originally the result
of the 2016 Winograd Schema Challenge
had the winner at 48\%. 
A person simply tossing a coin
would be expected to get 45\% as several
of the questions had more than 2 possible
answers. Unfortunately, 
the organisers had made an error in the input
file. When this was fixed, the winning entry
from Quan Liu of the University of Science and
Technology of China did 10\% better.}. This is  better 
than a person tossing a coin at random but
is still some way from the 90\% required
to win the cash prize. 

Another interesting alternative to the 
Turing test is the Lovelace 2.0 test \cite{lovelace2}. 
This is a refinement of the Lovelace Test 
of Bringsjord, Bello,
and Ferrucci in which
an  intelligent  system  must  originate  a  creative  concept
or  work  of  art
\cite{lovelace}. In the Lovelace 2.0
test, the player is asked to create some
artifact, like a story or a picture, 
that meets some constraints set
by a judge. For example, the
constraints might be to tell a story
in which ``a boy falls in love with a 
girl, the girl falls in love with the
boy's twin brother, the two twins
switch identities, but the girl now
realizes she loved the first twin all along''. 
A player passes the Lovelace 2.0 test if the created artifact
is considered by the judge to be as novel
and creative as an average unskilled
human would achieve. 
Passing such a test would require
several high level  cognitive  capabilities  including  
common-sense  reasoning, a  theory  of  mind, 
discourse  planning,   natural  language  processing,
and creativity.

We will not argue further for the advantages
or disadvantages of any of these alternatives.
We will, however, note that the 
change we propose to the Turing test,
in which we remove the asymmetry between
humans and machines, can be applied
to many of these tests. For instance,
we will  describe shortly how we
can modify the Lovelace 2.0 test to 
remove the asymmetry between humans
and machines in this test.

\section{The meta-Turing test}

We now turn to our contribution, which is a modification 
to the Turing and other similar tests of machine intelligence. 
Implicit in Turing's imitation game is the
assumption that it takes intelligence to
spot intelligence. Intelligent humans
are the judges. This introduces an
asymmetry into the test. Humans
alone judge the humans and machines
participating in the test. What
if we remove this asymmetry by
having all the humans and machines
judge each other? 

In the one-to-one version of the
meta-Turing test, a group
of humans and machines have 
pairwise conversations as in Turing's
imitation game. Each
participant then decides if the other
player in the conversation is human
or machine. 
In the one-to-two version of the
meta-Turing test, ever human and machine
in the test judge an imitation game
between every possible pair of humans
and machines. Each pair consists of one
human and one machine. 
The human or machine judging
each imitation game much decide which of
the pair is human and which is machine. 
A machine can be said to
pass such a meta-Turing test if it 
is consistently mistaken for human by the 
humans taking the test {\bf and} 
it reliably can identify the
machines recognised 
by humans to be machines. 
A meta-Turing test is thus a series
of conversations in which each agent
is trying to work out which of the
other agents are machine or human. 
An agent playing this test can no
longer just try to deceive. The agent
must actively try to work out which
of the other agents are human and which
are machine. 

Note that to pass the test, we do not ask 
a machine to recognise reliably any machines that 
are themselves potentially passing the
meta-Turing test. We cannot expect
a machine to distinguish apart
humans from machines that are being
consistently mistaken for human. 
Without this restriction, the meta-Turing test would
be oddly non-monotonic. We
could replace a program in a meta-Turing
test by a more capable program and some other program
that passed the test might no longer pass. 

There are other details of the test we
have not yet specified. How many players
should there be in the test? How long
should the conversations be? How do we
define reliably and consistently? We will get
to suggestions for how to determine
these details shortly. 
Another issue
is who should play this game. For simplicity,
we might suppose we have an equal number
of humans and machines (but this is not 
necessary). Including machines that we know
are poor at the test is also problematic. It may give
an advantage to the other machines. 
For instance, we might deliberately submit multiple programs
to the test, many of which we know are easily
recognised to try to weight the game to our advantage.
Therefore we might decide that only the best programs
currently available can play the game. In addition, we might
not run all pairwise conversations but limit
them to those where there is no conflict of 
interest. For example, two programs submitted
by the same author or by authors with a 
professional relationship might be deemed
to represent a conflict of interest. There
is a risk otherwise that programs might
benefit by collusion. 

Finally, once the first machine passes the
meta-Turing test, there is an argument 
that we should never run the test again. 
Before this time, the test measures if
a machine can consistently pass for
human and can reliably itself differentiate
between human and machine.
After we have such machines, 
we can no longer say that we can 
reliably differentiate between human
and machine. 

\section{Passing the meta-Turing test}

One of the problems with the Turing test
is that is rewards deceptive behaviour. 
Could the meta-Turing test
not be similarly deceived? For the sake of
argument, suppose we have a soft bot
even more deceptive than Eugene Goodman
that can consistently pass for human
by deception. 
We might decide simply to add a simple routine to
this soft bot that mechanically runs 
a Winograd Schema Challenge in every
conversation it has. With this routine, the soft bot
might reliably be able to tell humans apart
from machines that are not mistaken for
human. 
However, any human taking the meta-Turing
test would hopefully spot this
simple trick and no longer consider
the soft bot as human. 
In general, for a machine to pass a meta-Turing
test, it needs to {\em both} ask and answer
questions in a way that is responsive and
human like. We discuss shortly additional refinements
that will further hinder spoofing.

\section{The inverted Turing test}

Watt has proposed the related ``inverted Turing test''
\cite{invertedturingtest}. 
In an inverted Turing test, a machine has to distinguish 
as well between humans and machines as humans
can. In Watt's own words,
\begin{quote}
{\em
``a system passes if it is itself unable to distinguish between two
humans, or between a human and a machine that can pass the normal
Turing test, but which can discriminate between a human and a machine
that can be told apart by a normal Turing test with a human
observer.''
}
\end{quote}
The inverted Turing test maintains an asymmetry
between humans and machines as only the
machines are doing the judging. 
A meta-Turing test is roughly speaking
the combination of a Turing test and an inverted
Turing test. Watt claimed that the idea of
the inverted Turing test was, however, not
meant to be a replacement of the original Turing test. Instead,
he proposed that it be seen more as a thought 
experiment than as a goal for AI research. 
He argued that it might provide insight
into human psychology, ``other minds'' and
related philosophical issues. He suggested that
it adds something that is well hidden in 
the original Turing test. 

As he and others have recognised \cite{frenchturingtest}, 
it would be easy to 
cheat an inverted Turing test. We could, 
for instance, simply write a computer program
that simply administers a Winograd Schema Challenge. 
The meta-Turing test counters this problem,
requiring the machine to both appear
intelligent and to recognise intelligence. 
The two sided nature of the meta-Turing test 
guard against
the weaknesses of either
side: simple chat bots that deceive
Turing's original test, or mechanical
testing programs that defeat
tests like the inverted Turing test. 

\section{The reverse Turing test}

Another related but different test is the reverse
Turing test. 
In a reverse Turing test, we reverse
(some of) the roles of humans and machines.
One form of reverse Turing test is
when a human player tries
to trick a human judge into thinking
that they are a computer. 
Another form of reverse Turing is 
a CAPTCHA where a computer judge
rather than a human tries to decide
if a player is a human or a computer \cite{captcha}.
In both these types of reverse Turing test, we still 
have an asymmetry between humans
and computers. In the former,
only humans are doing the judging,
whilst in the latter, 
only computers are doing the judging. 
A meta-Turing test is therefore
different to a reverse Turing test
as there is no asymmetry between humans 
and computers in the former. 

\section{Some refinements}

We now propose some further refinements
of the meta-Turing test. In fact, many of
these refinements can be applied to the
original Turing test itself.

\subsection{College educated adult rule}

Turing talks about playing the imitation
game with an adult human \cite{turingmind}.
There's a strong argument then that
playing the imitation game with a
machine pretending to be a 13 year old 
Ukrainian boy violates the requirements
of the test. However, we might strengthen
Turing's (somewhat implicit)
requirement further and insist that participants
are college educated adults or machines
trying to imitate them. 

\subsection{Domain choice rule}

Turing discusses a conversation with 
an unrestricted domain \cite{turingmind}. 
This gives chat bots the opportunity to focus
on whimsical conversations that have proved
likely to deceive judges. We 
might counter this with a refinement
that limits the domain. For example, 
we might have an outside judge provide
a topic for conversation at regular
intervals. As a second example, we 
might divide the conversation into 
two halves, and have each player choose
the topic of conversation within their half.
The other player would be required
to follow the topic or risk failing
the test. 

\subsection{Test duration rule}

As explained earlier, Turing did not
provide a concrete recommendation
for the length of the game, though
some have interpreted his remarks
to suggest judges have just 
5 minutes of conversation with a player in which
to make up their minds \cite{turingmind}.
Results with Turing style tests
suggests 5 minutes is just too short. 
We might therefore consider, say,  
longer 30 minute conversations. 
Alternatively, we might consider
an open test, where each player
continues the test until they are
certain whether the other player
is human or machine. 

\subsection{Success rule}

Turing also 
did not provide a concrete recommendation
for identifying when a machine passed the imitation
game, though
some have interpreted his remarks
to suggest that the machine needed
to mis-recognised as human by 30 percent
of the judges \cite{turingmind}.
This appears to be too low a bar. 
Ultimately we would like machines
to be unrecognisable apart from 
humans. This would translate into
a judging rule as follows. 
In an one-to-two
Turing test, a machine passes when it is recognised
as human 50 percent of the time.
In an one-to-two
meta-Turing test, we would additionally
require that the machine recognises correctly the human
100 percent of the time when the other 
player in the pair is a machine that itself
is not mistaken as human by human judges. 
In an one-to-one Turing test,
a machine passes when no human judge identifies the
machine as a machine. In an one-to-one meta-Turing
test, we would additionally require that
the machine only identifies a machine as human
if that machine is mistaken
by the human judges as human. 
We might also require 
that the machine does not identify any human
as a machine. 

Unlike the previous 30 percent rule, 
these criteria might prove to be a little
too tough. Humans are not themselves
100 percent accurate. 
For instance, as mentioned earlier, in an online 
experiment on Mechanical Turk humans
only achieved 92 percent accuracy on Winograd 
Schema Challenge questions \cite{benderbaseline}.
It may therefore be appropriate to relax
these rules modestly. For instance, it might
be acceptable to ask merely that 
90 percent of human judges mistake
the machine as human in an one-to-one
meta-Turing test. Recall previously 
that this was set at 100 percent. 
Similarly, we might only require 
that machine identifies 
humans correctly in 90 percent 
of one-to-two meta-Turing test
when the player in the pair that is a machine 
is itself not mistaken as human by human judges. 
Again recall that this was previously set
at 100 percent. 

\subsection{Pool size rule}

We have not yet specified how many humans
and machines should be tested in a meta-Turing
test. To preserve the symmetry, we might demand
an equal number of humans and machines. 
Clearly, one of each is too small. Any human
would know without testing that they must 
be the human in the test. Two humans and two
machines would prevent this default assumption.
Nevertheless, it is probably too small a pool to
ensure accuracy. The final round of the most recent Loebner
prize has 4 humans and 4 machines. This is 
still likely too small a pool to produce
any sort of accuracy. The pool might
ideally have 
at least one dozen humans and one dozen machines.

\section{The meta-Lovelace test}

Let us return to the Lovelace 2.0 test.
Like the Turing test, this has an inherent
asymmetry between the humans and machines. 
The meta-Lovelace 2.0 test removes this
asymmetry. We again have a group of
humans and machines. We run a sequence of 
Lovelace 2.0 tests between pairs in this group.
Each test is run in both directions, with
one player setting and judging the task,
and the other creating the artwork. 
Each participant then decides if the other
player in the test is human
or machine. As before,
a machine can be said to
pass the meta-Lovelace 2.0 test if it 
is consistently mistaken for human by the 
humans taking the test {\bf and} 
it reliably can identify the
machines recognised 
by humans to be machines. 
Such a test requires the machine
to have new skills. For instance, it
must have the appropriate natural
language and vision skills to judge
originality in a written or visual artwork. 

\section{The meta-Winograd Schema Challenge}

We can adapt the Winograd Schema challenge
in a similar way. In a meta-Winograd Schema Challenge,
a machine takes a Winograd Schema Challenge,
as well as invents and performs a Winograd Schema Challenge
on the other players. 
To pass a meta-Winograd Schema Challenge,
the machine needs to answer the Winograd Schema
Challenges set by the other players
accurately, and to set a Winograd
Schema Challenge that reliably differentiates between
humans and machines that fail to pass
their Winograd Schema
Challenges. We might decide, for instance, 
that accurately answering a Winograd Schema
Challenge requires 90\% or greater correctness.
Similarly, we might decide that reliably
differentiating between humans and machines
requires setting a test on which humans
get 90\% or greater accuracy and machines
which fail the Winograd Schema Challenges
get less than 90\%. 

One issue is that a machine might simply 
set a Winograd Schema Challenge by picking
at random a set of questions stored in a large
database. To prevent this, the judges
might provide a sequence of noun phrases, adjectives
or verbs, and require
that the questions use each in turn.
Setting a Winograd Schema Challenge will
then require some creativity. 
For instance, we might give a pair of noun
phrases like ``toy'' and ``grass'' and require 
a question
that uses these two noun phrases.
See Figure 3 for an example.
We might alternatively
give a pair of adjectives 
like ``short'' and ``tall''
and require that the question use them. 
\begin{figure}[htb]
{%
    \parbox{0.45\textwidth}{

{\em The toy was lost in the grass because it was short. What is short?}

0:  the toy, 1:     the grass

\vspace{1em}

{\em The toy was lost in the grass because it was tall. What is too tall?}

0:  the toy, 1:     the grass
}}

\caption{Example of a pair of Winograd Schema Challenge questions
  invented for the noun phrases ``toy'' and ``grass''.}
\end{figure}

\section{Peer grading}

The meta-Turing test is related to the task
of peer grading. In a peer grading exercise, we
have a group of agents, each of whom does a
task (e.g. writing an essay, or answering some exam 
questions) and grades some subset of the other agents at
that task. We can consider the meta-Turing test
as a peer grading exercise in which the task
being peer graded is imitating a human. 
In \cite{wecai2014}, a fixed point equation
is proposed for constructing peer grades which
is a sum of two terms, the first being a weighted
sum of the grades given to an agent, the weights 
being the (estimated) grades of the agents doing
the grading, and a penalty term for mis-grading
the other agents. 
The idea of weighting the sum is that
we want to favour the opinion of the
agents who do well on the test. 
We might considering adapting
such a fixed point equation to the meta-Turing
test. However, there is one significant difference
since in a meta-Turing test, we actually 
know the ground truth. We know precisely
who is human and who is not. In a peer grading
exercise, the ground truth is unknown. 
Nevertheless, in a meta-Turing test, we 
also want to combine two similar terms: the
estimates of the other agents about whether 
you are human or not, and your ability to 
estimate correctly whether the other
agents are human or not.

\section{Discussion}

Turing side-stepped the original philosophical
question of whether a machine can really think.
The meta-Turing test also side-steps this
question. It merely determines if the machine
is behaving in a way that requires thinking
in humans. Namely, can the machine produce intelligent
conversation like a human {\em and} recognise 
intelligent conversation. Of course, people like Searle 
with his famous Chinese Room thought
experiment argue that it is possible to
get the observable behaviour right without
having the associated mental states \cite{chineseroom}.
There are, however, numerous arguments
against Searle's objections. For example, the 
Systems Reply argues that the system as whole 
understands Chinese \cite{levesqueturing}. More
generally, Turing was more interested
in an operational perspective. AI can be 
seen to have succeeded when we can no longer tell it
apart from human behaviour. We do not 
need our machines to be actually thinking. It
is good enough that they can do whatever
intelligent tasks we ask of them. 

As machines intrude more and more into
our lives, there are concerns
about whether machines will be
intentionally or unintentionally
mistaken for humans. For instance,
the recently proposed ``Turing Red Flag'' law
requires that machines should not be designed
to be mistaken for human and
that machines should announce themselves as
machines at the start of any interaction to
avoid confusion
\cite{redflag}. Such a law would
require autonomous
vehicles to be clearly identified
as such, so that other drivers do not mistake them
as driven by humans. We may even have
special lanes where only autonomous vehicles will
be allowed. 
Turing's imitation game explicitly 
challenges the idea that machines 
should not be designed to be mistaken
for human. The meta-Turing 
test does not change this. 
We may therefore require machines
to be exempted from such a rule
to permit their intelligence to be tested.

\section{Conclusions}

We have proposed an alternative to the Turing
test to tackle some of the criticisms made of 
Turing's original imitation game. In a meta-Turing
test, we remove the asymmetry between humans
and machines. Both humans and machines
judge who appears human and who appears
to be a machine. This calls upon an
assumption implicit in Turing's original
proposal that it takes intelligence to 
identify intelligence. To pass a meta-Turing test,
a computer needs both to be consistently
mistaken for human and for the computer to reliably 
recognise apart machines from humans. We
also propose some refinements like the length
of the test, and the rule for passing the test. 
The meta-Turing test
is more difficult to pass
than the Turing test. In a logical
sense, this is trivially the case
as a meta-Turing test includes
within it a Turing test of passing
for human conversation, plus the
additional requirement of identifying
apart other machines from humans. 
However, it also will defeat simply
deceptive tricks used currently by
chat bots. A computer will not, for instance, be
able just to pretend to be whimsical
or a non-native speaker. It will also actively
have to decide if the other player is
human or computer. 

\section{Acknowledgements}

Toby Walsh receives support from the 
European Research Council via an Advanced
Research grant, and the Asian Office of Aerospace Research
and Development. 
Data61 (formerly NICTA) is supported
by the Australian Government through the Department of Communications and the Australian Research Council through the ICT Centre of Excellence Program. 
\bibliographystyle{aaai}


\end{document}